%% file: main.tex
\documentclass{article} 
\PassOptionsToPackage{dvipsnames}{xcolor}
\usepackage{iclr2026_conference,times}

\input{preamble}

\usepackage[utf8]{inputenc} 
\usepackage[T1]{fontenc}    
\usepackage{url}            
\usepackage{amsfonts}       
\usepackage{microtype}      

\definecolor{mydarkblue}{rgb}{0,0.08,0.45}
\usepackage[breaklinks,colorlinks,allcolors=mydarkblue]{hyperref}
\hypersetup{
  pdftitle={StreamTTT: Reconciling Real-Time Perception and Long-Term Memory in Streaming VLMs},
  pdfauthor={Joya Chen, Zeyun Zhong, Mike Zheng Shou},
  pdfsubject={Preprint}
}

\title{\ours: Reconciling Real-Time Perception and Long-Term Memory in Streaming VLMs}

\author{%
  Joya Chen$^{1}$ \qquad
  Zeyun Zhong$^{2}$ \qquad
  Mike Zheng Shou$^{1}$\thanks{Corresponding author.} \\
  $^{1}$National University of Singapore \\
  $^{2}$Karlsruhe Institute of Technology
}

\iclrfinalcopy

\makeatletter
\newcommand{\unmarkedfootnote}[1]{%
  \begingroup
  \renewcommand{\thefootnote}{}%
  \renewcommand{\@makefnmark}{}%
  \footnote{#1}%
  \endgroup
  \addtocounter{footnote}{-1}%
}
\makeatother

\begin{document}
\maketitle
\unmarkedfootnote{This work is still a work in progress.}
\lhead{Preprint}

\input{sec/0_abstract}
\input{sec/1_intro}
\input{sec/2_related}
\input{sec/3_method}
\input{sec/4_results}
\input{sec/5_conclusion}

\section*{Ethics Statement}
This work introduces no new human-subject data collection. All training and
evaluation data are drawn from existing video and multimodal datasets and are
used subject to their stated licenses and terms. We do not intentionally
collect, infer, or release personally identifiable information. Because some
source datasets contain recordings of people and everyday activities,
downstream use should respect the privacy, consent, and usage conditions
established by those datasets.
We disclose that large language models (LLMs) were used only for minor copy-editing and language polishing (grammar and phrasing). LLMs played no role in ideation, algorithm design, experiments, data analysis, or results. All scientific claims and artifacts are the original work of the authors.

\section*{Reproducibility statement}
The paper describes the model architecture (Sec.~\ref{sec:method}), training-data construction (Sec.~\ref{sec:data}), and reported evaluation tracks. Ablation studies (Sec.~\ref{sec:ablation}) clarify component contributions. Appendix~\ref{sec:implementation_details_supp} specifies the memory branch and temporal-windowing implementation.

\bibliography{main}
\bibliographystyle{iclr2026_conference}

\input{sec/A_implementation}
\input{sec/B_data}

\end{document}

%% file: preamble.tex
\usepackage{xcolor}
\usepackage{colortbl}
\usepackage{graphicx}
\usepackage{amsmath}
\usepackage{booktabs}
\usepackage{multirow}
\usepackage{xspace}
\usepackage{amssymb}
\usepackage{algorithm}
\usepackage{algpseudocode}
\usepackage{bm}
\usepackage{adjustbox}
\usepackage{enumitem}
\usepackage{wrapfig}
\usepackage{placeins}
\usepackage{tikz}
\usetikzlibrary{positioning,arrows.meta,calc,fit,backgrounds,shapes.geometric}

\algrenewcommand{\algorithmiccomment}[1]{{\footnotesize \hfill // #1}}

\newcommand{\ours}{StreamTTT\xspace}
\definecolor{ourblue}{RGB}{232, 240, 254}
\newcommand{\best}[1]{%
  \begingroup
  \setlength{\fboxsep}{1pt}%
  \colorbox{yellow!35}{\textbf{#1}}%
  \endgroup}

%% file: sec/0_abstract.tex
\begin{abstract}
Humans effortlessly perceive the present while remembering the past, yet
streaming VLMs often trade off real-time perception against long-term memory.
Prior work shows that shortening the context can sharpen current-scene
perception at the expense of long-range recall. To reconcile these abilities,
we introduce \ours, which writes long-range history into online-updated fast
weights outside the attention context. This leaves a short sliding key-value
cache dedicated to recent evidence, mitigating attention dilution. We train
\ours\ jointly on offline long-video QA and a newly constructed real-time QA
corpus. On OVO-Bench, under each model's reported input protocol, \ours-4B
outperforms the same-scale SimpleStream-4B by $0.6$ points in real-time
perception and $5.3$ points in backward tracing. It also surpasses the larger
SimpleStream-8B by $0.73$ points on StreamingBench's Real-Time Visual
Understanding (RTVU) subset. Our code is publicly available at
\url{https://github.com/zeyun-zhong/StreamTTT}.
\end{abstract}

%% file: sec/1_intro.tex
\section{Introduction}
\label{sec:intro}

Like JARVIS in Marvel's \emph{Iron Man}, an ideal streaming video assistant
should understand the present, remember the past, and provide timely
assistance in real-world scenarios. Toward this vision, streaming VLMs now
support online dialogue~\citep{chen2024videollm,huang2025ovbench,
xiong2025streamchat}, continuous
commentary~\citep{chen2025livecc,xu2026streamingvlm,zhong2026flownar}, proactive
response~\citep{chen2024videollm,qian2025dispider,azad2026streamready,
zhang2026querystream,lu2026aura}, and native multimodal
interaction~\citep{thinkingmachines2026interaction}. These advances are
reflected in online video understanding
benchmarks~\citep{niu2025ovo,huang2025ovbench,lin2026streamingbench,
shi2026river}, which highlight real-time perception and long-term memory as two
central capabilities.

\begin{wrapfigure}[16]{r}{0.47\linewidth}
  \vspace{-4mm}
  \centering
  \includegraphics[width=\linewidth]{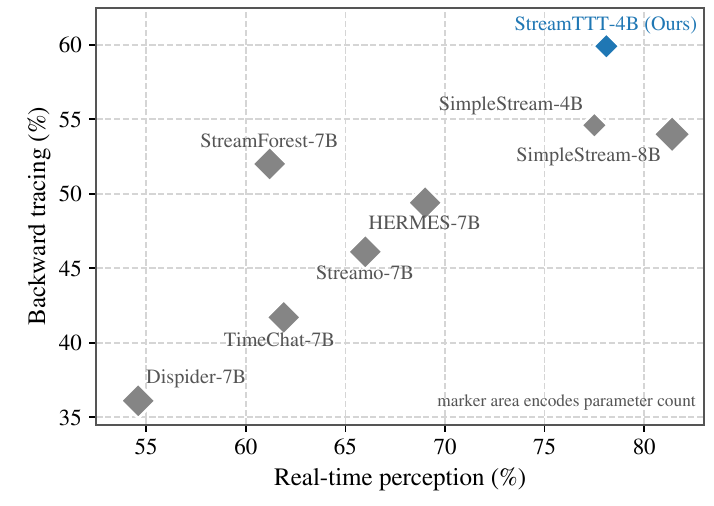}
  \vspace{-7mm}
  \caption{\textbf{OVO-Bench results}~\citep{niu2025ovo}. Marker area denotes
  model size.}
  \label{fig:pareto}
  \vspace{-2mm}
\end{wrapfigure}
Despite this progress, \citet{shen2026simple} reveal a
\emph{perception--memory trade-off}. Their recency baseline, which retains only
a short window of recent frames, outperforms more elaborate streaming systems on
real-time perception, while adding historical context can improve recall but
often weakens current-scene perception. This motivates separating recent
context from long-range state. However, most existing memory methods preserve
history through compression, retrieval, or token
merging~\citep{yang2025streammem,zhang2026hermes,di2025streaming,
zeng2025streamforest,chen2026streamingtom,xiao2026mukv,
ge2026selectstream,fan2026folio}, but ultimately feed selected history back
into the attention context. Injected history then consumes context capacity
and may dilute attention over recent evidence~\citep{shen2026simple,
liu2024lost}.

To realize this principle, we introduce \ours, a streaming VLM that stores
long-range history in online-updated fast
weights~\citep{sun2024learning,behrouz2025titans,zhang2025test,
zhong2026e2ttt,sun2026videosalmonns,liu2026spatialttt} outside the attention
context. A gated test-time training (TTT) branch realizes this memory alongside
self-attention, whose short sliding KV cache~\citep{xiao2024streamingllm}
remains dedicated to recent context. This separation preserves recent evidence
without expanding the attention context. To enable \ours\ to learn both
abilities, we jointly train it on offline long-video
QA~\citep{zhang2024llavavideo} and a real-time QA corpus built from existing
streaming and video datasets~\citep{xia2026streaming,grauman2022ego4d,
pan2023aria,krishna2017densecaptioning}. Together, the dual-memory architecture
and joint supervision allow \ours\ to preserve real-time perception while
recovering long-range history.

On OVO-Bench~\citep{niu2025ovo}, \ours-4B achieves $69.00$ averaged over
real-time perception and backward tracing. Compared with
SimpleStream-4B~\citep{shen2026simple}, it improves real-time perception by
$0.6$ points ($78.1$ vs.\ $77.5$) and backward tracing by $5.3$ points
($59.9$ vs.\ $54.6$). At matched parameter count under each paper's reported
input protocol, these results indicate stronger recall alongside current-scene
perception. On
StreamingBench~\citep{lin2026streamingbench}, we evaluate the Real-Time Visual
Understanding (RTVU) subset; \ours-4B scores $81.32$, exceeding
SimpleStream-8B ($80.59$) by $0.73$ points at half the parameter count.

In summary, our contributions are:
\begin{itemize}[leftmargin=1.4em,labelsep=0.4em,itemsep=3pt,topsep=2pt,parsep=0pt]
  \item \textbf{Reconciling real-time perception and long-term memory.}
  \ours\ separates the two demands: a short sliding KV cache preserves recent
  evidence for accurate real-time perception, while a parallel TTT branch
  stores long-range history outside the attention context for recall. This
  separation avoids displacing or diluting recent context.
  \item \textbf{A data construction strategy for both abilities.} We build a
  $112.4$K real-time QA corpus by relocating proactive queries to their answer
  time and adding action, spatial-reasoning, and captioning supervision.
  Combining this corpus with offline long-video QA provides complementary
  training for real-time perception and long-range recall.
  \item \textbf{Strong streaming results.} \ours-4B reaches $69.00$ on
  OVO-Bench and $81.32$ on StreamingBench RTVU. Against SimpleStream-4B,
  under each model's reported input protocol, it improves real-time perception
  by $0.6$ points and backward tracing by $5.3$ points.
\end{itemize}

%% file: sec/2_related.tex
\section{Related Work}
\label{sec:related}

\paragraph{Test-time training.}
Fast-weight memory treats rapidly updated parameters as sequence
state~\citep{schmidhuber1992fastweight,ba2016fastweights,
munkhdalai2017metanetworks}. Related recurrent mechanisms carry information
across context boundaries through segment recurrence, compressive memory,
recurrent tokens, or fast-weight linear attention
\citep{dai2019transformerxl,rae2020compressive,bulatov2022rmt,
katharopoulos2020transformers,schlag2021linear,irie2021recurrentfwp}.
Test-time training originated as self-supervised adaptation under distribution
shift~\citep{sun2020ttt} and was later reformulated as a sequence layer whose
online-trained model is the recurrent state~\citep{sun2024learning}. Titans
and chunkwise TTT improve memory dynamics and efficiency
\citep{behrouz2025titans,zhang2025test,zhong2026e2ttt}. Video applications
include long-form generation~\citep{dalal2025one}, streaming audio-visual
memory in video-SALMONN S~\citep{sun2026videosalmonns}, and long-horizon
spatial reasoning in Spatial-TTT~\citep{liu2026spatialttt}. These works focus
on generation or domain-specific memory, whereas \ours\ targets the general
problem of balancing real-time perception and long-range recall in streaming
VLMs.

\paragraph{Streaming VLM memory.}
Streaming VLMs extend earlier video memory and token-reduction techniques
\citep{wu2019longterm,wu2022memvit,he2024malmm,song2024moviechat,
ryoo2021tokenlearner,bolya2023tome} to a continually growing observed history.
Existing systems retain token summaries or event structures
\citep{qian2024videostreaming,zhang2025flash,zeng2025streamforest,
fan2026folio}, manage historical KV states
\citep{di2025streaming,yang2025streammem,zhang2026hermes,
kim2025infinipot,ning2026livevlm,chen2026streamingtom}, or retrieve
hierarchical and on-demand memories
\citep{ge2026selectstream,xie2026fluxmem,liang2026oasis,
xie2026streamrag}. StreamingVLM~\citep{xu2026streamingvlm} retains
attention sinks with asymmetric recent vision and text windows. Because these
approaches generally return history to backbone attention, the injected
history can compete with recent evidence~\citep{shen2026simple}.
SelectStream~\citep{ge2026selectstream} and FOLIO~\citep{fan2026folio}
mitigate this through selective retrieval; \ours\ instead stores history in a
parallel fast-weight state, leaving the KV cache for recent context.

\paragraph{Streaming VLM datasets.}
Training corpora for streaming VLMs span three main interaction formats.
VideoLLM-online~\citep{chen2024videollm},
VideoChat-Online~\citep{huang2025ovbench},
StreamChat~\citep{xiong2025streamchat}, and
ProVideLLM~\citep{chatterjee2025providellm} use temporally aligned dialogue or
procedural assistance. Supervision for answer timing appears in
Dispider~\citep{qian2025dispider},
Streamo~\citep{xia2026streaming}, QueryStream, StreamReady, and Response-G1
\citep{zhang2026querystream,azad2026streamready,ma2026responseg1}.
StreamBridge, LiveStar, and StreamMind provide proactive-response
supervision~\citep{wang2025streambridge,yang2025livestar,ding2025streammind},
whereas LiveCC~\citep{chen2025livecc} densely interleaves video and speech for
continuous commentary. These corpora emphasize dialogue and response timing.
We instead construct real-time perception supervision by moving each proactive
query to its first answerable frame, then pair it with offline long-video QA
for recall.

\paragraph{Streaming VLM benchmarks.}
OVO-Bench~\citep{niu2025ovo}, OVBench~\citep{huang2025ovbench}, and
StreamingBench~\citep{lin2026streamingbench} evaluate observed video prefixes
through real-time perception, retrospective recall, or proactive response.
RIVER~\citep{shi2026river} further structures questions by temporal demand,
while SVBench, OmniMMI, LiViBench, and PhoStream extend evaluation to
multi-turn, livestream, or mobile audio-visual settings
\citep{yang2025svbench,wang2025omnimmi,wang2026livibench,lu2026phostream}.
Complementary offline benchmarks expose the complete video and test reasoning
over minutes to days~\citep{wu2024longvideobench,zhou2025mlvu,
wang2025lvbench,yang2025egolife}. We use both regimes to assess current
perception and recall after evidence leaves the recent window.

%% file: sec/3_method.tex
\section{Method}
\label{sec:method}

A streaming VLM must preserve recent evidence for real-time perception while
retaining older information for later recall. Existing systems often route both
through the same finite attention context, where historical tokens can displace
or dilute recent evidence. \ours\ separates these roles: sliding-window
attention models recent context, while an online-trained fast-weight state
stores a compressed trace of earlier context outside attention.

\subsection{Fast-Weight Memory with Large-Chunk TTT}
\label{sec:background}

Test-time training (TTT) treats the parameters $\mathbf W$ of a small model
$f_{\mathbf W}$ as recurrent fast-weight state
\citep{sun2024learning,zhang2025test,zhong2026e2ttt}. Given an input token,
a TTT layer forms a query, a key, and a value $(\bm q_t,\bm k_t,\bm v_t)$. The
standard TTT write operation minimizes a self-supervised key--value
reconstruction loss and updates the fast weights online:
\begin{equation}
  \ell_t(\mathbf W)
  = \frac{1}{2}\big\| f_{\mathbf W}(\bm k_t)-\bm v_t \big\|_2^2,
  \qquad
  \mathbf W_t
  = \mathbf W_{t-1}
  - \eta_t
  \left.\nabla_{\mathbf W}\ell_t(\mathbf W)\right|_{\mathbf W=\mathbf W_{t-1}} .
  \label{eq:ttt-write}
\end{equation}
The updated model is read in the following query:
\begin{equation}
  \bm r_t = f_{\mathbf W_t}(\bm q_t).
  \label{eq:ttt-read}
\end{equation}
The fast weights thereby summarize previous key--value associations without
explicitly retaining all past tokens. Direct tokenwise updates are sequential.
Large-Chunk TTT (LaCT)~\citep{zhang2025test} instead partitions the sequence
into chunks of $C$ tokens and aggregates their weighted losses
into a single update, where chunk~$j$ spans tokens $(j-1)C+1, \dots, jC$:
\begin{equation}
  \bm g_j
  =
  \left.
  \nabla_{\mathbf W}
  \sum_{t=(j-1)C+1}^{jC}
  \eta_t\,\ell_t(\mathbf W)
  \right|_{\mathbf W=\mathbf W_{j-1}},
  \qquad
  \mathbf W_j
  =
  \operatorname{Update}(\mathbf W_{j-1},\bm g_j).
  \label{eq:lact-update}
\end{equation}
LaCT decouples the apply and update operations. For causal streaming, we retain
its apply-then-update order: every query in chunk~$j$ is read with
$\mathbf W_{j-1}$, and $\mathbf W_j$ is exposed only to later chunks.
Sliding-window attention handles within-chunk interactions, while fast weights
carry cross-chunk history. However, our branch does not use LaCT's
chunk-averaged momentum rule. We instantiate $\operatorname{Update}$ with the
closed-form E$^2$-TTT transition~\citep{zhong2026e2ttt}: all per-token
gradients in a chunk are evaluated at the shared chunk-start weights, while
token-dependent learning rate, momentum, and decay coefficients are retained
by two scalar kernels. Before implementation stabilizers, the transition
exactly recovers the corresponding chunk-end fast weights and momentum without
materializing token-indexed matrix states. The released layer additionally
clips the two aggregated updates and normalizes fast weights at chunk
boundaries. Appendix~\ref{sec:app-branch} gives the complete equations,
reconstruction target, and recurrent buffers.

\input{imgs/fig_method}

\subsection{Overview}
\label{sec:overview}

We process a video as ordered temporal windows
$\{\mathbf X_i\}_{i=1}^{N}$ (Fig.~\ref{fig:method}). Every decoder layer carries
two complementary memories: a sliding key--value (KV) cache
$\mathcal K$ for recent context, and a recurrent TTT state
$\mathcal S$ that compresses information from earlier windows.
The two branches are fused by a learnable gate. Across windows, both memories
are carried forward while globally continuous positions preserve temporal
order. Their sizes do not grow with elapsed video length, although the
fixed-size TTT state remains a lossy summary whose capacity is evaluated in
\S\ref{sec:ablation}.
We index tokens by $t$, TTT chunks of $C$ tokens by $j$, and temporal windows
by $i$. A window spans one or more chunks, and both memories are carried across
all three levels.

\subsection{Parallel TTT Memory Branch}
\label{sec:arch}

We build on Qwen3-VL and convert each self-attention block into a
\emph{hybrid} layer that retains the pretrained attention path and adds a
parallel TTT branch. For layer input $\bm x_t$, sliding-window attention (SWA)
reads recent KV states, while TTT reads and updates its recurrent state:
\begin{equation}
  \big(\bm{o}_t^{\text{SWA}},\, \mathcal{K}_t\big)
    = \mathrm{SWA}\big(\bm{x}_t \,;\, \mathcal{K}_{t-1}\big),
    \qquad
  \big(\bm{o}_t^{\text{TTT}},\, \mathcal{S}_t\big)
    = \mathrm{TTT}\big(\bm{x}_t \,;\, \mathcal{S}_{t-1}\big),
    \label{eq:branches}
\end{equation}
where $\mathcal{K}_t$ is the sliding cache of recent keys and values,
and $\mathcal{S}_t$ is the fixed-size recurrent state. Its
central component is the fast weights $\mathbf W_t$, which carry the compressed history through write and read operations in
Eqs.~\eqref{eq:ttt-write}--\eqref{eq:ttt-read}. Alongside them, $\mathcal{S}_t$
holds the small, bounded auxiliary quantities needed to resume the update rule
at a forward-pass boundary. These preserve the convolutional receptive field
and global TTT-chunk alignment when the stream is cut into multiple forwards;
the small numerical stabilizers around the closed-form update are specified in
Appendix~\ref{sec:app-branch}. All components are of fixed size, so
$|\mathcal{S}_t|$ is constant in the number of processed tokens.
Because $\mathcal{S}_t$ remains outside the attention context, long-range recall does not consume slots from the recent window. We evaluate this placement at matched
budget in \S\ref{sec:ablation} (Table~\ref{tab:abl_memory_type}).
The branch outputs are fused through a learnable channel-wise gate
$\bm\alpha\in\mathbb R^d$:
\begin{equation}
  \bm{o}_t
    = \bm{o}_t^{\text{SWA}}
    + \tanh(\bm\alpha)\odot\bm{o}_t^{\text{TTT}}.
    \label{eq:fuse}
\end{equation}
The gate is initialized near zero
\citep{alayrac2022flamingo,zhang2024llamaadapter,dalal2025one}, keeping the
initial function close to the pretrained attention path while the model learns
to use long-term memory.

\subsection{Streaming Inference over Temporal Windows}
\label{sec:streaming}

\paragraph{Temporal windowing.}
We partition the sampled frames into $N$ contiguous windows by wall-clock time.
Given a target window duration $\Delta$, the nominal window $i$ contains frames
whose timestamps fall in $[(i{-}1)\Delta,\, i\Delta)$; its right boundary is
then advanced by fewer than one temporal patch so that every window is
compatible with the visual encoder. The encoder maps the aligned frames to a
token block $\mathbf{X}_i$. The textual prefix and QA suffix are placed in the
first and final windows, respectively (App.~\ref{sec:app-window}).

\paragraph{Dual memory and sequential forward.}
We feed windows in order, carrying both memories across steps. After each
window, the KV cache is pruned to its most recent $L$ tokens, whereas the recurrent
TTT state $\mathcal S$ is carried without eviction as a fixed-size
compressed history. The final window produces the answer logits
(Alg.~\ref{alg:multiforward}).
We denote the initialized per-layer TTT state before the first window by
$\mathcal S_{\mathrm{init}}$.

\paragraph{Globally continuous M-RoPE.}
The M-RoPE indexer~\citep{wang2024qwen2vl,bai2025qwen3vl} resets its cursor
for each window, which would reuse positions across the stream. We preserve
each window's locally computed 3D positions and add a running scalar offset:
\begin{equation}
  \bm{p}_i \;=\; \bm{p}_i^{\text{loc}} + (m_{i-1}+1),
  \qquad
  m_i \;=\; \max\big(\bm{p}_i\big),
  \qquad m_0 = -1,
  \label{eq:rope}
\end{equation}
where $m_i$ is the running maximum over all position axes. This preserves
intra-window spatial-temporal structure while matching the contiguous
positions of a single full-video pass.

\begin{algorithm}[t]
\caption{Multi-Forward Streaming Inference over Temporal Windows}
\label{alg:multiforward}
\begin{algorithmic}[1]
\Require windows $\{\mathbf{X}_i\}_{i=1}^{N}$ with per-window video grids; attention span $L$
\Ensure next-token logits $\hat{\bm{y}}$ at the QA position
\State $\mathcal{K} \gets \varnothing$;\quad $\mathcal{S} \gets \mathcal S_{\mathrm{init}}$;\quad $m \gets -1$ \Comment{KV cache, recurrent state, running pos.\ max}
\For{$i = 1$ \textbf{to} $N$}
  \State $\bm{p}_i^{\text{loc}} \gets \textsc{M-RoPE-Index}(\mathbf{X}_i)$ \Comment{recompute window positions from a zero cursor}
  \State $\bm{p}_i \gets \bm{p}_i^{\text{loc}} + (m + 1)$ \Comment{add scalar offset to all axes, Eq.~\eqref{eq:rope}}
  \State $m \gets \max(\bm{p}_i)$ \Comment{single max over the $(t,h,w)$ axes and sequence}
  \vspace{2pt}
  \State $(\hat{\bm{y}},\, \widetilde{\mathcal K},\, \mathcal{S}) \gets
         \text{LLM}(\mathbf{X}_i, \, \bm{p}_i, \, \mathcal{K},\, \mathcal{S})$
         \Comment{forward through both memories}
  \State $\mathcal{K} \gets \mathrm{prune}_{L}(\widetilde{\mathcal K})$ \Comment{keep most recent $L$ tokens}
\EndFor
\State \Return $\hat{\bm{y}}$ \Comment{logits from the final (QA) window}
\end{algorithmic}
\end{algorithm}

%% file: imgs/fig_method.tex
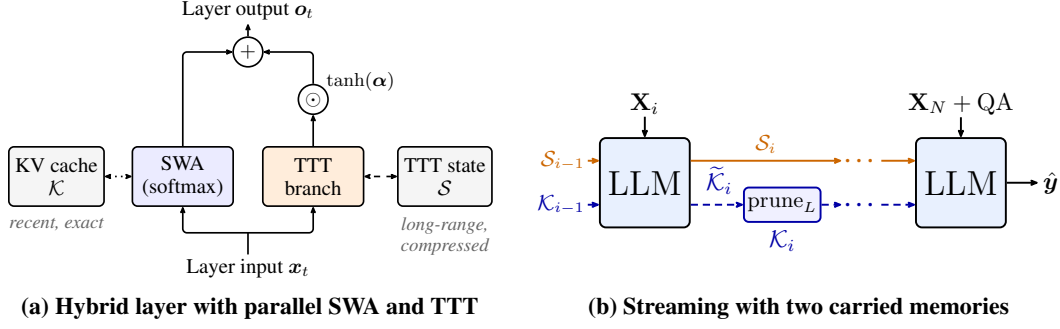
\begin{figure}[t!]
\centering
\tikzset{
  box/.style={rounded corners=2.5pt, draw, thick, align=center, inner sep=3pt,
              minimum height=10mm, minimum width=17mm, font=\normalsize},
  attn/.style={box, fill=blue!8},
  ttt/.style={box, fill=orange!14},
  mem/.style={box, fill=black!4, minimum width=15mm},
  op/.style={draw, thick, circle, inner sep=0pt, minimum size=5mm},
  io/.style={align=center, inner sep=2pt, font=\normalsize},
  ann/.style={font=\small, align=center, inner sep=1pt},
  flbl/.style={font=\small\itshape, text=black!60, align=center,
               inner sep=1pt},
  wire/.style={->, thick, rounded corners=2pt},
  rd/.style={->, thick, dotted},
  stream model/.style={rounded corners=2.5pt, draw, thick, fill=ourblue,
                       minimum width=12mm, minimum height=13mm, font=\Large},
  stream input/.style={font=\normalsize, align=center, inner sep=1pt},
  stateflow/.style={->, thick, draw=orange!80!black},
  kvflow/.style={->, thick, densely dashed, draw=blue!65!black},
  statelabel/.style={font=\small, text=orange!80!black, inner sep=1pt},
  kvlabel/.style={font=\small, text=blue!65!black, inner sep=1pt},
  prune/.style={rounded corners=2pt, draw=blue!65!black, thick,
                fill=blue!6, minimum width=9mm, minimum height=5mm,
                font=\small, align=center, inner sep=1.5pt},
}

\begin{minipage}[b]{0.46\textwidth}
\centering
\adjustbox{max width=\linewidth}{%
\begin{tikzpicture}[
  font=\normalsize,
  >={Stealth[round,length=4.6pt,width=3.5pt]},
]
  \node[attn] (attn) at (-1.10,1.45) {SWA\\(softmax)};
  \node[ttt]  (ttt)  at ( 1.10,1.45) {TTT\\branch};
  \node[mem,left =0.45cm of attn] (kv) {KV cache\\$\mathcal{K}$};
  \node[mem,right=0.55cm of ttt]  (st) {TTT state\\$\mathcal{S}$};
  \node[op]   (gate) at ( 1.10,2.75) {$\odot$};
  \node[op]   (sum)  at ( 0   ,3.55) {$+$};
  \node[io]   (out)  at ( 0   ,4.25) {Layer output $\bm o_t$};
  \node[io]   (h)    at ( 0   ,-0.10) {Layer input $\bm x_t$};

  \coordinate (fan) at (0,0.50);
  \draw[wire,-] (h.north) -- (fan);
  \draw[wire] (fan) -| (attn.south);
  \draw[wire] (fan) -| (ttt.south);

  \draw[<->,thick,dotted] (kv) -- (attn);
  \draw[<->,thick,dashed] (st) -- (ttt);

  \draw[wire] (attn.north) |- (sum.west);
  \draw[wire] (ttt.north) -- (gate.south);
  \draw[wire] (gate.north) |- (sum.east);
  \draw[wire] (sum.north) -- (out);
  \node[ann,anchor=west] at (1.28,3.05) {$\tanh(\bm\alpha)$};

  \node[flbl,below=0.14cm of kv] {recent, exact};
  \node[flbl,below=0.14cm of st] {long-range,\\compressed};
\end{tikzpicture}%
}
\par\vspace{2pt}
{\small\bfseries (a) Hybrid layer with parallel SWA and TTT\par}
\end{minipage}%
\hfill
\begin{minipage}[b]{0.50\textwidth}
\centering
\adjustbox{max width=\linewidth}{%
\begin{tikzpicture}[
  font=\normalsize,
  >={Stealth[round,length=4.6pt,width=3.5pt]},
]
  \node[stream model] (mi) at (1.55,1.15) {$\mathrm{LLM}$};
  \node[stream model] (mN) at (6.15,1.15) {$\mathrm{LLM}$};
  \node[stream input] (xi) at (1.55,2.30) {$\mathbf X_i$};
  \node[stream input] (xN) at (6.15,2.30) {$\mathbf X_N+\mathrm{QA}$};
  \draw[wire] (xi) -- (mi.north);
  \draw[wire] (xN) -- (mN.north);

  \coordinate (mis-in)  at ($(mi.west)+(0,0.32)$);
  \coordinate (mis-out) at ($(mi.east)+(0,0.32)$);
  \coordinate (mik-in)  at ($(mi.west)+(0,-0.32)$);
  \coordinate (mik-out) at ($(mi.east)+(0,-0.32)$);
  \coordinate (mNs-in)  at ($(mN.west)+(0,0.32)$);
  \coordinate (mNk-in)  at ($(mN.west)+(0,-0.32)$);

  \node[statelabel,anchor=east] (sprev) at (0.72,1.47)
    {$\mathcal S_{i-1}$};
  \node[kvlabel,anchor=east] (kprev) at (0.72,0.83)
    {$\mathcal{K}_{i-1}$};
  \draw[stateflow] (sprev.east) -- (mis-in);
  \draw[kvflow] (kprev.east) -- (mik-in);

  \node[statelabel] (sdots) at (4.70,1.47) {$\boldsymbol{\cdots}$};
  \draw[stateflow] (mis-out) --
    node[statelabel,above=2pt] {$\mathcal S_i$} (sdots.west);
  \draw[stateflow] (sdots.east) -- (mNs-in);

  \node[prune] (prune) at (3.55,0.83) {$\mathrm{prune}_L$};
  \node[kvlabel] (kdots) at (4.70,0.83) {$\boldsymbol{\cdots}$};
  \draw[kvflow] (mik-out) --
    node[kvlabel,font=\normalsize,above=3pt,pos=0.55]
    {$\widetilde{\mathcal K}_i$} (prune);
  \draw[kvflow] (prune) -- (kdots.west);
  \node[kvlabel,font=\normalsize,anchor=north,yshift=-2pt] at (prune.south)
    {$\mathcal{K}_i$};
  \draw[kvflow] (kdots.east) -- (mNk-in);

  \node[io,right=0.45cm of mN] (y) {$\hat{\bm{y}}$};
  \draw[wire] (mN) -- (y);
\end{tikzpicture}%
}
\par\vspace{0.45cm}
{\small\bfseries (b) Streaming with two carried memories\par}
\end{minipage}
\caption{\textbf{\ours overview.}
\textbf{(a)} At token $t$, the pretrained sliding-window attention branch maps
$(\bm x_t,\mathcal K_{t-1})$ to
$(\bm o_t^{\mathrm{SWA}},\mathcal K_t)$, while the parallel TTT branch maps
$(\bm x_t,\mathcal S_{t-1})$ to
$(\bm o_t^{\mathrm{TTT}},\mathcal S_t)$. The bounded cache $\mathcal K$ stores
recent KV pairs inside the attention context, whereas the fixed-size state
$\mathcal S$ stores compressed history outside it. Their outputs are fused by
the channel-wise gate $\tanh(\bm\alpha)$, initialized near zero.
\textbf{(b)} Here $i$ indexes temporal windows. One LLM forward consumes
$\mathbf X_i$ and the two memories left by the preceding window. After the
token/chunk updates within that forward, $\mathcal S_i$ is carried without
eviction, while the resulting cache $\widetilde{\mathcal K}_i$ is pruned to
$\mathcal K_i$, its most recent $L$ tokens, before the next window. M-RoPE
positions remain globally continuous. The textual prefix and QA suffix enter
the first and final windows, respectively; the final forward yields next-token
logits $\hat{\bm y}$.}
\label{fig:method}
\end{figure}

%% file: sec/4_results.tex
\section{Results}

\subsection{Training Data for Real-Time Perception and Backward Tracing}
\label{sec:data}

Addressing the perception--memory trade-off requires supervision for both
temporal regimes (\S\ref{sec:intro}): the model must learn to interpret the
current scene while retaining information needed by later queries. Long-range
recall supervision is readily available from long-video QA corpora. We
therefore sample $119$K offline whole-video QA pairs from the long
($2$--$3$\,min) subset of LLaVA-Video-178K~\citep{zhang2024llavavideo}.
Real-time perception supervision is scarcer because existing streaming
corpora primarily target proactive response timing. To fill this gap, we build
a $112.4$K real-time QA corpus and combine it with the offline sample. The
resulting mixture is nearly balanced; \S\ref{sec:ablation} evaluates the
contribution of each half.

Streamo~\citep{xia2026streaming} reformulates existing video corpora as
large-scale proactive QA. A question is issued at time $t_q$, strictly before
the answer-relevant event ends at $t_a$, so the model must wait and respond only
after the necessary evidence arrives. This setup teaches \emph{when} to
answer, but does not specifically supervise perception when the evidence first
becomes available. We instead relocate each query to its answer time,
$t_q := t_a$. The converted query is thus posed as soon as its answer becomes
available and can be answered without waiting for future evidence.

We apply two filters to keep this supervision temporally local. Because many
Streamo items originate from temporal-grounding annotations, their answers are
tied to labeled segments rather than single frames. We retain only segments
shorter than $10$\,s, concentrating the supporting evidence near $t_q$, and
restrict source videos to $45$--$180$\,s to match the training horizon.
Applying this conversion to Streamo's refactored corpora---LLaVA-Video,
QVHighlights~\citep{lei2021qvhighlights},
EgoTimeQA~\citep{di2024groundvqa}, ActivityNet
Captions~\citep{krishna2017densecaptioning}, and
HowToCaption~\citep{shvetsova2024howtocaption}---produces the $80.1$K
converted block in Table~\ref{tab:data}.

Because the converted data inherit their source distribution, action and
spatial reasoning remain underrepresented. We add $32.3$K constructed or
repurposed examples whose labels come directly from ground-truth annotations
rather than model-generated answers. We select action- and spatial-centric
questions from EgoTimeQA~\citep{di2024groundvqa}. We also construct
anticipation questions from Ego4D Short-Term Anticipation
~\citep{grauman2022ego4d}; by construction, the target hand--object interaction
is not yet visible at query time. For spatial reasoning, we generate
four-option questions from the 2D/3D geometry of the Aria Digital
Twin~\citep{pan2023aria}. These questions cover image-plane direction,
inter-object layout, and relative distance, and each is answerable from a
single query frame. A small captioning split repurposed from ActivityNet
completes the $32.3$K block in Table~\ref{tab:data}. Appendix~\ref{sec:app-data}
details the full corpus composition, offline sampling, answerability and
visibility filters, question families, leakage-free timing, deduplication, and
distractor sampling.

\subsection{Comparison with Streaming Baselines}
\input{tables/main}

\ours-4B achieves a two-track average of $69.00$ on OVO-Bench, outperforming
HERMES-7B ($59.20$) by $9.80$ points and SimpleStream-8B
~\citep{shen2026simple} ($67.70$) by $1.30$ points, despite using half as many
parameters as the latter. The same-parameter-scale comparison with
SimpleStream-4B follows each system's reported input protocol rather than a
shared inference configuration: SimpleStream-4B uses its highest-average
reported window (16 frames at $1$\,fps), whereas \ours-4B uses $2$\,fps with a
4K-token sliding KV cache. Under these source-specific protocols, \ours-4B
raises the two-track average from $66.06$ to $69.00$, real-time perception from
$77.5$ to $78.1$, and the official Backward Tracing average from $54.6$ to
$59.9$. SimpleStream does not report the 4B per-task scores needed to compute
the episodic-recall metric
$\mathrm{ER}=(\mathrm{EPM}+\mathrm{ASI})/2$, so we use ER only for our
controlled ablations below. Under the stated protocols, the comparison
therefore shows higher backward-tracing accuracy without lower current-scene
accuracy. Larger
concurrent systems report higher absolute OVO-Bench scores through selective
latent or semantic memory
~\citep{ge2026selectstream,fan2026folio}; our result isolates the effectiveness
of separating recent context from long-range state at the $4$B scale.

On the StreamingBench RTVU subset, \ours-4B achieves $81.32$, exceeding
HERMES-7B ($79.44$) by $1.88$ points and SimpleStream-8B ($80.59$)
by $0.73$ points with half as many parameters as the latter.

\subsection{Ablation Studies}
\label{sec:ablation}

\paragraph{Ablation protocol.}
For all ablation studies, we use the Qwen3-VL-4B backbone with the same
optimizer, schedule, number of steps, and random seed across trained cells.
Each condition trains on the full data pool for its supervision sources
($119$K offline, $112.4$K real-time, and their union for joint).
Following \citet{shen2026simple}, we evaluate every cell on OVO-Bench's
Real-Time Visual Perception and Backward Tracing categories at $2$\,fps,
falling back to a video's native frame rate where it is below $2$\,fps, using
the same windowed-inference protocol. Because HLD primarily measures
hallucination robustness rather than episodic event recall, we report
$\mathrm{ER}=(\mathrm{EPM}+\mathrm{ASI})/2$ as the recall metric. We omit
StreamingBench from the ablations to reduce compute. The final model in
Table~\ref{tab:main_ovo} uses the full training recipe.

\paragraph{Architecture $\times$ data (Table~\ref{tab:abl_arch_data}).}
The architecture and data axes are complementary. With joint supervision,
disabling the TTT branch reduces the real-time average from $78.05$ to $68.19$
and ER from $59.50$ to $49.58$. With the full hybrid architecture,
offline-only training improves ER over the KV-only condition but lowers
real-time perception to $65.08$, below the frozen recency reference
($78.66$). Real-time-only training largely preserves current perception
($78.15$) but yields lower ER than joint training ($57.00$ vs.\ $59.50$).
Among the trained variants, the hybrid architecture with joint supervision
attains the highest ER and the highest mean of real-time perception and ER.
Its real-time score is $0.10$ points below real-time-only training, while its
ER is $2.50$ points higher, yielding the strongest overall balance.

\paragraph{Complete memory configurations at matched inference footprint
(Table~\ref{tab:abl_memory_type}).}
We compare complete memory configurations at a matched inference-time memory
footprint. All variants use an identical $4$K sliding KV cache as short-range
memory. For StreamMem~\citep{yang2025streammem} and
HERMES~\citep{zhang2026hermes}, we add a long-range KV store to the trained
sliding-KV checkpoint at inference, without method-specific retraining, and
match its physical memory to that of the fast weights. The fast-weight row
instead uses the jointly trained hybrid checkpoint. Under this protocol, the
fast-weight configuration reaches $78.05$ on real-time perception and $59.50$
on ER, outperforming both heuristic configurations. Because the checkpoints
and training procedures differ, this is a memory-matched but not
training-matched comparison: it evaluates the complete configurations rather
than isolating the memory mechanism alone.

\input{tables/ablation_arch_data}

\input{tables/ablation_memory_type}

\begin{figure}[t!]
  \centering
  \includegraphics[width=\textwidth]{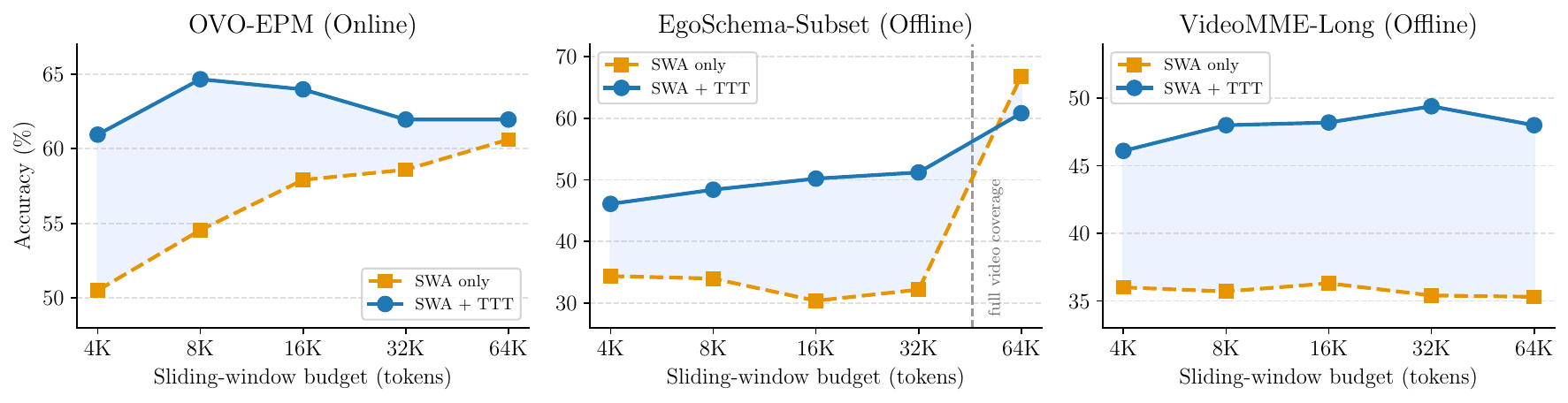}
  \caption{\textbf{Window-budget analysis.} Accuracy with and without the TTT
  state as the sliding attention window grows from $4$K to $64$K tokens.}
  \label{fig:capacity}
\end{figure}

\paragraph{Window-budget analysis (Fig.~\ref{fig:capacity}).}
Prior work varies memory-bank or visual-window size to characterize
bounded-context behavior, and recent diagnostics measure sensitivity to frame
budgets~\citep{he2024malmm,xu2026streamingvlm,tian2026memorybudget}. We adapt
this analysis to ask how much recent attention context the learned state can
replace. Holding all other settings fixed, we vary the sliding-window budget
from $4$K to $64$K tokens, with and without the TTT branch. We evaluate one
online episodic-memory track, OVO-EPM~\citep{niu2025ovo}, and two offline
long-video settings, EgoSchema-Subset~\citep{mangalam2023egoschema} and
VideoMME-Long~\citep{fu2025videomme}.

Across bounded $4$K--$32$K windows, the TTT state improves EgoSchema by
$11.7$--$19.8$ points and VideoMME-Long by approximately $10$--$14$ points.
On VideoMME-Long, sliding-window attention alone remains near $36$ because the
relevant evidence often falls outside the window. The fraction of missing
context recovered by TTT, however, depends on the recall demand. We quantify
this at a $4$K operating budget relative to the $64$K sliding-window reference.
On OVO-EPM, increasing the window from $4$K to $64$K raises the sliding-window
baseline from $50.51$ to $60.61$, a $10.10$-point gap. Adding TTT at $4$K
raises accuracy to $60.94$, slightly exceeding the $64$K sliding-window
reference by $0.33$ points. On
EgoSchema, the sliding-window baseline rises from $34.4$ at $4$K to $66.8$ at
$64$K, a $32.4$-point gap. TTT raises the $4$K result to $46.1$, recovering
$11.7$ points, or approximately $36\%$ of the gap. At $64$K, where the
EgoSchema window covers the full short clip, the TTT configuration scores
$60.8$ versus $66.8$ for sliding-window attention alone, a $6.0$-point
deficit. The fixed-size state
therefore recovers substantial missing context under bounded attention, but it
complements rather than replaces full-video attention when the entire video
fits in context.

\FloatBarrier

%% file: tables/main.tex
\begin{table*}[t!]
\centering
\caption{\textbf{Streaming benchmark results.}
Following SimpleStream~\citep{shen2026simple}, we report OVO-Bench
~\citep{niu2025ovo} Real-Time Visual Perception and Backward Tracing task
accuracies, category means, and their mean, together with StreamingBench
~\citep{lin2026streamingbench} RTVU accuracy. Input settings follow their
sources: SimpleStream-8B/4B use their best-average windows (4/16 frames at
$1$\,fps), while \ours-4B processes video at $2$\,fps with a 4K-token
sliding KV cache. ``--'' is unreported and $\dagger$ denotes Qwen2.5-VL-7B
with HERMES (4K tokens). Task abbreviations follow OVO-Bench. Yellow bold and
underlined entries are best and second-best; blue marks our model.}
\label{tab:main_ovo}
\renewcommand{\arraystretch}{1.1}
\setlength{\tabcolsep}{3pt}
\resizebox{\textwidth}{!}{%
\begin{tabular}{lc|c|cccccc|c|ccc|c|c}
\toprule
\multirow{3}{*}{Model} & \multirow{3}{*}{\shortstack{Input\\Setting}}
& \multirow{3}{*}{\shortstack{StreamingBench\\RTVU}}
& \multicolumn{12}{c}{\textit{OVO-Bench}} \\
\cmidrule(lr){4-15}
& &
& \multicolumn{7}{c|}{\textit{Real-Time Visual Perception}}
& \multicolumn{4}{c|}{\textit{Backward Tracing}}
& \multirow{2}{*}{Avg.} \\
\cmidrule(lr){4-10} \cmidrule(lr){11-14}
& & & OCR & ACR & ATR & STU & FPD & OJR & Avg.
& EPM & ASI & HLD & Avg. & \\
\midrule
Human
& -- & 91.46
& 94.0 & 92.6 & 94.8 & 92.7 & 91.1 & 94.0 & 93.2
& 92.6 & 93.0 & 91.4 & 92.3
& 92.77 \\
\midrule
\multicolumn{15}{c}{\textit{Offline Video LLMs}} \\
\midrule
Qwen2.5-VL-7B
& 1\,fps & 73.31
& 67.8 & 55.1 & 67.2 & 42.1 & 66.3 & 60.9 & 59.9
& 51.5 & 58.8 & 23.7 & 44.7
& 52.28 \\
LLaVA-OneVision-7B
& 32 & 71.12
& 66.4 & 57.8 & 73.3 & 53.4 & 71.3 & 62.0 & 64.0
& 54.2 & 55.4 & 21.5 & 43.7
& 53.85 \\
InternVL2-8B
& 16 & 63.72
& 67.1 & 60.6 & 63.8 & 46.1 & 68.3 & 56.5 & 60.4
& 48.2 & 57.4 & 24.7 & 43.4
& 51.90 \\
LLaVA-Video-7B
& 64 & --
& 69.1 & 58.7 & 68.8 & 49.4 & 74.3 & 59.8 & 63.4
& 56.2 & 57.4 &  7.5 & 40.4
& 51.86 \\
Qwen2-VL-7B
& 64 & 69.04
& 69.1 & 53.2 & 63.8 & 50.6 & 66.3 & 60.9 & 60.7
& 44.4 & \best{66.9} & 34.4 & 48.6
& 54.62 \\
LongVU-7B
& 1\,fps & --
& 55.7 & 49.5 & 59.5 & 48.3 & 68.3 & 63.0 & 57.4
& 43.1 & \underline{66.2} &  9.1 & 39.5
& 48.45 \\
\midrule
\multicolumn{15}{c}{\textit{Online / Streaming VLMs}} \\
\midrule
VideoLLM-online-8B
& 2\,fps & 35.99
&  8.1 & 23.9 & 12.1 & 14.0 & 45.5 & 21.2 & 20.8
& 22.2 & 18.8 & 12.2 & 17.7
& 19.26 \\
Flash-VStream-7B
& 1\,fps & 23.23
& 24.2 & 29.4 & 28.5 & 33.7 & 25.7 & 28.8 & 28.4
& 39.1 & 37.2 &  5.9 & 27.4
& 27.90 \\
Dispider-7B
& 1\,fps & 67.63
& 57.7 & 49.5 & 62.1 & 44.9 & 61.4 & 51.6 & 54.6
& 48.5 & 55.4 &  4.3 & 36.1
& 45.35 \\
TimeChat-Online-7B
& 1\,fps & 75.28
& 75.2 & 46.8 & 70.7 & 47.8 & 69.3 & 61.4 & 61.9
& 55.9 & 59.5 &  9.7 & 41.7
& 51.80 \\
StreamForest-7B
& 1\,fps & 77.26
& 68.5 & 53.2 & 71.6 & 47.8 & 65.4 & 60.9 & 61.2
& \underline{58.9} & 64.9 & 32.3 & 52.0
& 56.60 \\
Streamo-7B
& 1\,fps & --
& 79.2 & 57.8 & 75.0 & 49.4 & 64.4 & 70.1 & 66.0
& 54.6 & 52.0 & 31.7 & 46.1
& 56.05 \\
HERMES-7B$^\dagger$
& 1\,fps & 79.44
& 85.2 & 64.2 & 71.6 & 53.4 & 74.3 & 65.2 & 69.0
& 48.5 & 62.2 & 37.6 & 49.4
& 59.20 \\
SimpleStream-8B
& 4 & \underline{80.59}
& \best{94.0} & \best{85.3} & \best{82.8} & \best{65.7} & \underline{77.2} & \best{83.2} & \best{81.4}
& 51.9 & 58.1 & \underline{52.1} & 54.0
& \underline{67.70} \\
SimpleStream-4B & 16 & -- & \multicolumn{6}{c|}{--} & 77.5
& \multicolumn{3}{c|}{--} & \underline{54.6} & 66.06 \\
\rowcolor{ourblue} \ours-4B & 2\,fps / 4K
& \best{81.32} & \underline{91.3} & \underline{80.7} & \underline{80.2}
& \underline{56.2} & \best{82.2} & \underline{77.7} & \underline{78.1}
& \best{60.9} & 58.1 &  \best{60.8} & \best{59.9} & \best{69.00} \\

\bottomrule
\end{tabular}
}
\end{table*}

%% file: tables/ablation_arch_data.tex
\begin{table}[t]
\centering
\caption{\textbf{Memory architecture $\times$ training data.}
All trained cells use the Qwen3-VL-4B backbone, the same optimization schedule
and seed. Following \citet{shen2026simple}, evaluation covers OVO-Bench's
Real-Time Visual Perception and Backward Tracing categories at $2$\,fps.
``Sliding KV only'' disables the TTT branch while leaving the remaining
architecture unchanged. The frozen recency baseline
(\citealp{shen2026simple}-style, 4 frames) requires no training. The hybrid
model with joint supervision achieves the strongest balance across the two
tracks. Because HLD primarily measures hallucination robustness rather than
episodic event recall, we report
$\mathrm{ER}=(\mathrm{EPM}+\mathrm{ASI})/2$ as the recall metric.}
\label{tab:abl_arch_data}
\setlength{\tabcolsep}{6pt}
\begin{tabular}{ll|cc|c}
\toprule
Memory & Training data & RT Avg & ER & Avg. \\
\midrule
\multicolumn{5}{l}{\textit{Frozen reference (no training)}} \\
Recency window (4 frames) & --- & 78.66 & 51.25 & 64.96 \\
\midrule
Sliding KV only & joint & 68.19 & 49.58 & 58.89 \\
Hybrid (KV + TTT) & offline-only & 65.08 & 57.50 & 61.29 \\
Hybrid (KV + TTT) & realtime-only & \textbf{78.15} & 57.00 & 67.58 \\
\rowcolor{ourblue}
Hybrid (KV + TTT) & joint & 78.05 & \textbf{59.50} & \textbf{68.78} \\
\bottomrule
\end{tabular}
\end{table}

%% file: tables/ablation_memory_type.tex
\begin{table}[t]
\centering
\small
\caption{\textbf{Complete memory configurations at matched inference
footprint.}
All rows use the Qwen3-VL-4B backbone and joint training data, but they are not
training-matched. The heuristic rows add a long-range store to the trained
\emph{sliding-KV$+$joint} checkpoint at inference, without method-specific
retraining; the fast-weight row uses the jointly trained hybrid checkpoint.
The table holds the $4$K short-range KV cache fixed and matches each heuristic
store's storage footprint to that of the fast weights; it therefore compares
complete configurations rather than isolating the memory mechanism.}
\label{tab:abl_memory_type}
\setlength{\tabcolsep}{4pt}
\begin{tabular}{l|c|ccc}
\toprule
Long-range memory & Short-range memory  & RT Avg & ER & Avg. \\
\midrule
None (sliding KV only) & \multirow{4}{*}{KV cache (4K)} & 68.19 & 49.58 & 58.89 \\
StreamMem~\citep{yang2025streammem} &  & 75.48 & 55.82 & 65.65 \\
HERMES~\citep{zhang2026hermes} &  & 75.11 & 57.34 & 66.23 \\
\rowcolor{ourblue}
Fast weights (ours) & & \textbf{78.05} & \textbf{59.50} & \textbf{68.78}  \\
\bottomrule
\end{tabular}
\end{table}

%% file: sec/5_conclusion.tex
\section{Conclusion}

We address the perception--memory trade-off in streaming video understanding
by separating recent context from long-range state. \ours\ combines a bounded
sliding KV cache with a parallel recurrent TTT memory and is jointly trained
for real-time perception and backward tracing. On OVO-Bench, \ours-4B
preserves the real-time perception of a same-scale recency baseline while
improving long-range recall; it also surpasses the larger SimpleStream-8B on
the StreamingBench RTVU subset. Our ablations show that the hybrid architecture and joint
supervision provide the strongest balance between the two capabilities. The
fixed-size recurrent state remains a lossy summary on recall-intensive videos,
however, and should complement rather than replace full attention when the
entire sequence fits in context.
Improving its capacity and selectivity is a promising direction for reliable
long-horizon streaming assistants.

%% file: sec/A_implementation.tex
\appendix

\section{Implementation Details}
\label{sec:implementation_details_supp}

This appendix specifies the memory branch, the recurrent state carried across
forward passes, and the temporal-windowing procedure deferred from
\S\ref{sec:background} and \S\ref{sec:streaming}.

\subsection{Memory-Branch Instantiation}
\label{sec:app-branch}

\paragraph{Memory heads and reconstruction target.}
The TTT branch contains multiple fast-weight heads. Omitting the head index,
each head parameterizes the gated MLP
\begin{equation}
  f_{\mathbf W}(\bm x)
  = \mathbf W^{(1)}\!\left(
  \mathrm{SiLU}(\mathbf W^{(0)}\bm x)
  \odot(\mathbf W^{(2)}\bm x)\right).
\end{equation}
A branch-specific projection followed by a causal depthwise convolution
produces $(\bm q_t,\bm k_t,\bm v_t)$, with $\ell_2$-normalized queries and
keys. Let $\mathcal N$ denote an affine LayerNorm learned with the rest of the
model. Following~\citet{dalal2025one}, we specialize the write objective in Eq.~\eqref{eq:ttt-write} to
\begin{equation}
  \widetilde{\bm v}_t=\mathcal N(\bm v_t-\bm k_t),
  \qquad
  \ell_t(\mathbf W)
  =\frac{1}{2}\big\|\mathcal N(f_{\mathbf W}(\bm k_t))
  -\widetilde{\bm v}_t\big\|_2^2,
\end{equation}
and use the residual readout
$\bm r_t=\bm q_t+\mathcal N(f_{\mathbf W}(\bm q_t))$.
The per-head residual readouts are concatenated and mapped by the branch output
projection to $\bm o_t^{\mathrm{TTT}}$ in Eq.~\eqref{eq:fuse}.

\paragraph{E$^2$-TTT coefficients and frozen-gradient recurrence.}
Consider one chunk and use local indices $t\in\{1,\ldots,C\}$, with
$(\mathbf W_0,\mathbf M_0)$ inherited from the preceding chunk. In the released
configuration, the learning-rate, momentum, and decay heads are head-wise. If
$a_t,b_t,c_t$ are their learned logits, the coefficients used by the code are
\begin{equation}
  \widehat\eta_t=\eta_{\mathrm{base}}\sigma(a_t),\qquad
  \eta_t=\widehat\eta_t/C,\qquad
  \beta_t=\sigma(b_t)^{1/16},\qquad
  \alpha_t=\alpha_{\mathrm{base}}\sigma(c_t),\qquad
  \gamma_t=1-\widehat\eta_t\alpha_t.
  \label{eq:app-ttt-coefficients}
\end{equation}
All reported experiments use $C=1024$,
$\eta_{\mathrm{base}}=10^{-4}$, and $\alpha_{\mathrm{base}}=0.1$.
Thus the decay factor is formed before the update coefficient is normalized by
the chunk size. Suppressing head and fast-weight-matrix indices, E$^2$-TTT
defines the following within-chunk recurrence~\citep{zhong2026e2ttt}:
\begin{equation}
  \mathbf G_t
  =-\left.\nabla_{\mathbf W}\ell_t(\mathbf W)
  \right|_{\mathbf W=\mathbf W_0},\qquad
  \mathbf M_t=\beta_t\mathbf M_{t-1}+\eta_t\mathbf G_t,
  \qquad
  \mathbf W_t=\gamma_t\mathbf W_{t-1}+\mathbf M_t.
  \label{eq:app-ttt-recurrence}
\end{equation}
The evaluation point $\mathbf W_0$ is shared by all $C$ gradients; using
$\mathbf W_{t-1}$ here would instead describe fully token-wise TTT and would
invalidate the parallel closed form. The intermediate states in
Eq.~\eqref{eq:app-ttt-recurrence} are analytical only: they are neither
materialized nor used for within-chunk readout, which uses $\mathbf W_0$.

\paragraph{Closed-form chunk transition.}
Define the suffix products and cumulative ratio
\begin{equation}
  \widetilde\beta_t=\prod_{i=t+1}^{C}\beta_i,\qquad
  \widetilde\gamma_t=\prod_{i=t+1}^{C}\gamma_i,\qquad
  R_t=\sum_{i=t}^{C}\frac{\widetilde\gamma_i}{\widetilde\beta_i},
  \label{eq:app-ttt-kernels}
\end{equation}
where an empty product equals one. Unrolling
Eq.~\eqref{eq:app-ttt-recurrence} then gives the exact pre-stabilization
chunk-end states
\begin{align}
  \mathbf M_C
  &=\widetilde\beta_0\mathbf M_0
    +\sum_{t=1}^{C}\eta_t\widetilde\beta_t\mathbf G_t,
  \label{eq:app-ttt-momentum-closed}\\
  \widehat{\mathbf W}_C
  &=\widetilde\gamma_0\mathbf W_0
    +\widetilde\beta_0R_1\mathbf M_0
    +\sum_{t=1}^{C}\eta_t\widetilde\beta_tR_t\mathbf G_t.
  \label{eq:app-ttt-weight-closed}
\end{align}
The momentum and weight updates therefore require distinct scalar kernels,
$\eta_t\widetilde\beta_t$ and
$\eta_t\widetilde\beta_tR_t$, respectively. They are evaluated by log-space
cumulative sums and reuse the same per-token activation gradients.

\paragraph{Implementation stabilizers.}
The released implementation clips each fast-weight matrix in each of the two
aggregated gradient terms to Frobenius norm at most one. After adding the carry
terms, it rescales every row of $\widehat{\mathbf W}_C$ to the corresponding
row norm recorded at the start of the forward call. It can also omit a leading
prefix whose momentum-kernel coefficients are below $10^{-9}$. Consequently,
the exact-equivalence statement above applies to the closed-form core; clipping,
post-chunk normalization, finite precision, and the negligible-prefix shortcut
are explicit numerical stabilizers around that core.

\paragraph{Per-layer recurrent state.}
Let $\mathbf W=(\mathbf W^{(0)},\mathbf W^{(1)},\mathbf W^{(2)})$ collect the three
fast-weight matrices of every memory head, and let $\mathbf M$ collect their
momentum states. Two auxiliary buffers preserve global chunk alignment across
the chunked, multi-pass execution of \S\ref{sec:streaming}. Up to the numerical
stabilizers described above, this gives the same closed-form updates as a
continuous pass. The first retains the last few projected tokens seen by the
depthwise convolution, so that its
receptive field carries across a pass boundary. The second handles chunk
boundaries: a forward pass may end before a chunk of $C$ tokens is complete,
and such a trailing partial chunk does not update $\mathbf W$ immediately.
Instead its per-token quantities
$(\bm{k},\bm{v},\bm{\eta},\log\bm{\gamma},\log\bm{\beta})$ are retained and
prepended to the next pass, so that the chunk produces its update once enough
new tokens arrive. The decay and momentum gates are kept in log space because
the chunkwise update accumulates their products along the chunk as cumulative
sums. Since this buffer holds at most $C-1$ tokens, chunk boundaries follow the
global token index and are unaffected by how the stream is cut into windows.
The per-layer state carried across windows is therefore
\begin{equation}
  \mathcal{S} \;=\; \big(\,
  \underbrace{\mathbf W}_{\text{fast weights}},\;
  \underbrace{\mathbf M}_{\text{momentum}},\;
  \text{convolution prefix},\;
  \text{partial-chunk buffer}
  \,\big),
\end{equation}
where every component has a fixed maximum size, so $|\mathcal{S}|$ is
independent of the number of tokens processed.

\subsection{Temporal Windowing}
\label{sec:app-window}

\paragraph{Patch-size alignment.}
Frames are assigned to windows by wall-clock time as in \S\ref{sec:streaming},
giving $N$ contiguous windows for a target duration $\Delta$. The visual encoder
additionally requires each window's frame count to be a multiple of its temporal
patch size $p$. We satisfy this by advancing the boundary \emph{index} rather
than by padding. Let $c_i$ be the number of sampled frames that fall strictly
before the $i$-th split time $i\Delta$. Window $i$ then ends at frame
$e_i=p\lceil c_i/p\rceil$, and window $i{+}1$ begins at frame $e_i+1$. With
$p=2$, for instance, a wall-clock boundary falling after frame $7$ is advanced to
frame $8$, so window $i$ takes one frame that by timestamp belongs to its
successor.

In general a window borrows at most $p-1$ frames from the next one, and the
windows remain a contiguous partition of the sampled frames. Since the frame
sampler already yields a multiple of $p$ frames in total, the trailing window is
aligned as well and no padding frame is ever inserted. The per-window visual
token blocks therefore concatenate into exactly the token layout of a single
full-video pass, with the same number of tokens in the same order. We require
$\Delta$ to be large enough that every wall-clock window contains at least $p$
sampled frames, which guarantees that alignment never empties a window.

\paragraph{Token layout across windows.}
The full prompt forms a single conceptual sequence
\begin{equation}
  \big[\,\underbrace{\text{system}+\text{instruction}}_{\text{prefix}}\,\big]
  \;\big[\,\underbrace{\mathbf X_1\,\mathbf X_2\,\cdots\,\mathbf X_N}_{\text{visual tokens}}\,\big]
  \;\big[\,\underbrace{\text{question}+\text{answer}}_{\text{QA suffix}}\,\big],
\end{equation}
where $\mathbf X_i$ denotes the visual tokens from window $i$. The first
forward pass contains the textual prefix and $\mathbf X_1$, intermediate
passes contain only their visual tokens, and the final pass contains
$\mathbf X_N$ followed by the QA suffix.
During training, the answer tokens are teacher-forced targets; during
inference, the question and assistant prefix are supplied in the final pass
and the answer is generated autoregressively. Generated answer tokens read the
post-prompt recurrent state but do not update the fast weights or the
partial-chunk buffer, preventing the memory from writing its own response.

%% file: sec/B_data.tex
\section{Data Construction Details}
\label{sec:app-data}

This appendix documents the training-corpus composition
(Table~\ref{tab:data}) and the construction procedures deferred from
\S\ref{sec:data}. We describe the offline sampling protocol, the
proactive-to-real-time conversion, and the two annotation-derived datasets:
Aria Digital Twin (ADT) spatial reasoning and future-action prediction (FAP).
For the generated datasets, labels are computed directly from ground-truth
annotations rather than synthesized by a model.

\input{tables/dataset}

\subsection{Offline Whole-Video Supervision}
\label{sec:app-data-offline}

The offline half of training is sampled from existing long-video QA rather
than newly constructed. We restrict the long-video subset of
LLaVA-Video-178K~\citep{zhang2024llavavideo} to clips lasting $2$--$3$ minutes
and randomly retain $50\%$, yielding approximately $119$K whole-video QA
pairs. Combining these with the $112.4$K real-time examples in
Table~\ref{tab:data} produces the near-balanced joint mixture used to train the
main model.

\subsection{Proactive-to-Real-Time Conversion}
\label{sec:app-data-realtime}

The $80.1$K upper block of Table~\ref{tab:data} is derived from Streamo's
proactive QA~\citep{xia2026streaming}. In the original construction, a
question arrives at $t_q$ before its answer-relevant event ends at $t_a$, and the
model must defer its response until the evidence becomes available. We convert
each item by setting $t_q:=t_a$, thereby posing the question at its annotated
answer time rather than before it. The conversion changes only the query
timestamp; the source question, answer, and answer options remain unchanged.
We apply the procedure to Streamo's versions of LLaVA-Video,
QVHighlights~\citep{lei2021qvhighlights},
EgoTimeQA~\citep{di2024groundvqa}, ActivityNet
Captions~\citep{krishna2017densecaptioning}, and
HowToCaption~\citep{shvetsova2024howtocaption}.

We apply two filters to keep the converted supervision temporally local. Many
Streamo items originate from temporal-grounding annotations and are associated
with a segment $[s,e]$ rather than a single frame. We retain only short
segments and restrict source videos to the $45$--$180$\,s duration range.
After filtering, the conversion yields the $80.1$K real-time block in
Table~\ref{tab:data}.

\subsection{QA Generation from ADT Annotations}
\label{sec:app-data-adt}

We construct spatial-reasoning QA directly from the ground-truth annotations
of the Aria Digital Twin dataset~\citep{pan2023aria}. ADT provides per-frame
2D object boxes, per-object oriented 3D boxes, object poses in a shared world
frame, and the device trajectory for each egocentric sequence. We use the RGB
stream and its unique annotation timestamps as the time axis. Each query is
anchored to one timestamp, and its label is computed from the corresponding
annotations.

\paragraph{Single-frame answerability.}
Every referenced object must be visible in the query frame. We therefore
discard questions about off-screen objects, including single-object egocentric
``behind me'' relations. An object at frame $f$ passes the visibility gate only
if its 2D box covers more than $2\%$ of the image, its annotated visibility
exceeds $0.8$, and its center falls within $591$ pixels ($0.42\times1408$) of
the image center, which restricts objects to the un-vignetted central disk of
the fisheye RGB sensor. We retain an object as an \emph{anchor} only if it is
annotated \emph{static}, so that a single world pose is exact for the whole
sequence, and passes this gate in at least $30$ frames.

Object references are generated conservatively. We use the semantic category
when it identifies a unique instance in the scene, and otherwise use a
normalized instance name only when that name is unique. We discard structural
parts and names distinguished solely by an arbitrary index (e.g.\ ``cabinet
door A''), which may not be resolvable from the image.

\paragraph{Question families.}
The pipeline specifies the following three families of four-option
multiple-choice questions, with the correct option's verbatim text as the
answer and option order shuffled. Let $\mathbf c_u\in\mathbb R^3$ be the
world-frame center of object $u$, obtained by transforming its local 3D box by
the object pose, and let $\mathbf p_f$ be the ground-truth device position at
frame $f$.
\begin{itemize}
  \item \textbf{Ego-quadrant.} ``In my current view, where is the $X$?'' The
  answer is the image quadrant (\textsc{upper/lower}$\times$\textsc{left/right})
  containing the center of $X$'s 2D box, which we require to be at least $140$
  pixels ($10\%$ of the image width) from both image midlines so that boundary
  cases are excluded. This family uses only 2D boxes.
  \item \textbf{Object--object.} ``Where is $A$ relative to $B$ from my point of
  view?'' with four options
  $\{\textsc{front,back}\}\times\{\textsc{left,right}\}$. The horizontal
  relation is determined from the 2D box centers, which must differ by at least
  $120$ pixels. The depth relation is determined by the sign of
  $\lVert\mathbf c_A-\mathbf p_f\rVert-\lVert\mathbf c_B-\mathbf p_f\rVert$,
  whose magnitude must be at least $0.40$\,m.
  \item \textbf{Relative distance.} ``Among these objects, which is closest to
  $T$?'' The label is the minimum corner-to-corner distance between world-frame
  3D boxes. We retain an example only if the closest candidate beats the
  runner-up by a multiplicative margin of $1.3\times$ and if $T$ and all four
  candidates pass the visibility gate in the query frame.
\end{itemize}

To limit simple label priors, we balance answers within each family, cap the
number of questions per object and object pair, and prefer query frames that
maximize the minimum visibility among referenced objects. The resulting ADT
split contributes $6.8$K examples to Table~\ref{tab:data}.

\subsection{Future-Action Prediction (FAP) Data Construction}
\label{sec:app-data-fap}

We construct the Future-Action Prediction (FAP) training split from Ego4D
Short-Term Object Interaction Anticipation (STA) annotations~\citep{grauman2022ego4d}.
Given an egocentric clip truncated at observation time $t_{\mathrm{obs}}$, the
task is to predict the next annotated hand--object interaction as a (verb, noun)
pair selected from four options. We use STA rather than Long-Term Anticipation
because it provides an observation frame and a \emph{time-to-contact} (TTC) for
each interaction, so contact occurs at $t_{\mathrm{obs}}+\mathrm{TTC}$. We
retain examples with $\mathrm{TTC}\in[0.5,2.0]$\,s (mean $1.58$\,s), which keeps
the target outside the observed prefix while remaining close enough to be
anticipated from pre-contact evidence such as reaching motion and gaze.

\paragraph{Deduplication and disambiguation.}
STA may annotate the same action at multiple nearby observation frames with
decreasing TTC. We group candidates by (clip, verb, noun) and retain the frame
whose TTC is closest to the midpoint of the retained range, leaving at most one
example per action type in a clip. We then discard frames carrying multiple
simultaneous interaction labels, and restrict $t_{\mathrm{obs}}$ to
$[45,180]$\,s so that every question has substantial observed context.

\paragraph{Distractor sampling.}
We pair each ground-truth action with three distractors. Whenever possible, one
is a \emph{hard in-context negative}: an action annotated elsewhere in the same
clip, making it scene-plausible but incorrect at the query time. Such an option
is available for $95.9\%$ of the questions. The remaining distractors are drawn
from a global frequency-weighted pool of observed (verb, noun) pairs. The four
options must refer to distinct objects, and we exclude a distractor if its
object is interacted with within $\pm2$\,s of the target contact time, so that
no distractor is incidentally correct. The final option order is randomized.

\paragraph{Resulting corpus.}
The pipeline yields 10{,}613 FAP questions from 2{,}029
Ego4D clips. Observation times range from $45$ to $180$\,s, with a mean of
$107.9$\,s. Correct-option frequencies range from $24.4\%$ to $25.7\%$ across
the four positions, indicating no substantial marginal position imbalance.
For each question, the provenance sidecar records the source annotation UID,
TTC, contact time, ground-truth action, and selected distractors.

%% file: tables/dataset.tex
\begin{table}[ht]
\centering
\small
\begin{tabular}{lllr}
\toprule
\textbf{Split} & \textbf{Source} & \textbf{Task type} & \textbf{\# QA} \\
\midrule
\multicolumn{4}{l}{\emph{Converted from Streamo proactive QA ($t_q := t_a$)}} \\
LLaVA-Video         & LLaVA-Video & --                    & 32.0K \\
QVHighlights        & QVHighlights & --                   & 19.6K \\
EgoTimeQA           & EgoTimeQA   & --                    & 6.8K \\
ActivityNet         & ActivityNet & --                    & 9.0K \\
HowToCaption        & HowToCaption & Captioning           & 12.7K \\
\cmidrule(l){4-4}
\multicolumn{3}{r}{\emph{subtotal}} & 80.1K \\
\midrule
\multicolumn{4}{l}{\emph{Constructed / repurposed in this work}} \\
Caption-ActivityNet & ActivityNet & Captioning            & 3.0K \\
EgoTimeQA-act       & EgoTimeQA   & Action QA             & 10.8K \\
EgoTimeQA-spatial   & EgoTimeQA   & Spatial QA            & 1.1K \\
ADT$^{\star}$       & ADT         & Spatial reasoning     & 6.8K \\
FAP$^{\star}$       & Ego4D-STA   & Anticipation          & 10.6K \\
\cmidrule(l){4-4}
\multicolumn{3}{r}{\emph{subtotal}} & 32.3K \\
\midrule
\multicolumn{3}{r}{\textbf{Total}} & \textbf{112.4K} \\
\bottomrule
\end{tabular}
\caption{\textbf{Composition of the $112.4$K real-time training corpus.}
The upper block converts Streamo's proactive QA by moving each query to its
annotated answer time ($t_q:=t_a$; \S\ref{sec:app-data-realtime}). A dash
indicates that the source construction does not provide a task label. The
lower block contains examples constructed or repurposed in this work; stars
mark the two annotation-derived datasets detailed in
\S\ref{sec:app-data-adt} and \S\ref{sec:app-data-fap}. The separate $119$K
offline whole-video sample is not included.}
\label{tab:data}
\end{table}